\documentclass[journal,twoside,web]{ieeecolor}
\usepackage{array}
\usepackage[table]{xcolor}
\usepackage{uma_fall_detection}
\usepackage{cite}
\usepackage{amsmath,amssymb,amsfonts}
\usepackage{algorithmic}
\usepackage{graphicx}
\usepackage{textcomp}
\usepackage{wrapfig}
\usepackage{epstopdf}
\usepackage{support-caption}
\usepackage{subcaption}
\usepackage{tabularray}
\usepackage[T1]{fontenc}
\usepackage[utf8]{inputenc}
\usepackage{booktabs} 
\usepackage{changes}
\def\BibTeX{{\rm B\kern-.05em{\sc i\kern-.025em b}\kern-.08em
    T\kern-.1667em\lower.7ex\hbox{E}\kern-.125emX}}
\definecolor{abstractbg}{rgb}{0.89804,0.94510,0.83137}
\begin{document}
\title{Towards Privacy-Supporting Fall Detection via Deep Unsupervised RGB2Depth Adaptation}
\author{ Hejun Xiao$^{\star}$$^{2}$, Kunyu Peng$^{\star}$$^{1}$, Xiangsheng Huang$^{*}$$^{2}$ , Alina Roitberg$^{1}$, Hao Li$^{2}$,  Zhaohui Wang$^{2}$and Rainer Stiefelhagen$^{1}$
\thanks{The research leading to these results was supported by the National Key R\&D Program of China (Grant Nos. 2020YFC2006406) and the SmartAge project sponsored by the Carl Zeiss Stiftung (P2019-01-003; 2021-2026).
The authors would like to thank the consortium for the successful cooperation.
\textit{(Corresponding author: Xiangsheng Huang, the Professor in Xiong'an Institute of Innovation and the Institute of Automation, Chinese Academy of Science. The authors marked with $^{\star}$ are both first authors)}
$^{1}$Authors are with Institute for Anthropomatics and Robotics, Karlsruhe Institute of Technology, Germany. (E-mail: \{kunyu.peng, alina.roitberg, rainer.stiefelhagen\}@kit.edu).$^{2}$Authors are with Xiong'an Institute of Innovation, Chinese Academy of Science, China. (E-mail: \{xiaohejun, huangxiangsheng, lihao\}@xii.ac.cn, wangzhaohui0311@gmail.com)}}

\IEEEtitleabstractindextext{%
\fcolorbox{abstractbg}{abstractbg}{%
\begin{minipage}{\textwidth}%
\begin{wrapfigure}[23]{r}{3.1in}%
\includegraphics[width=3in]{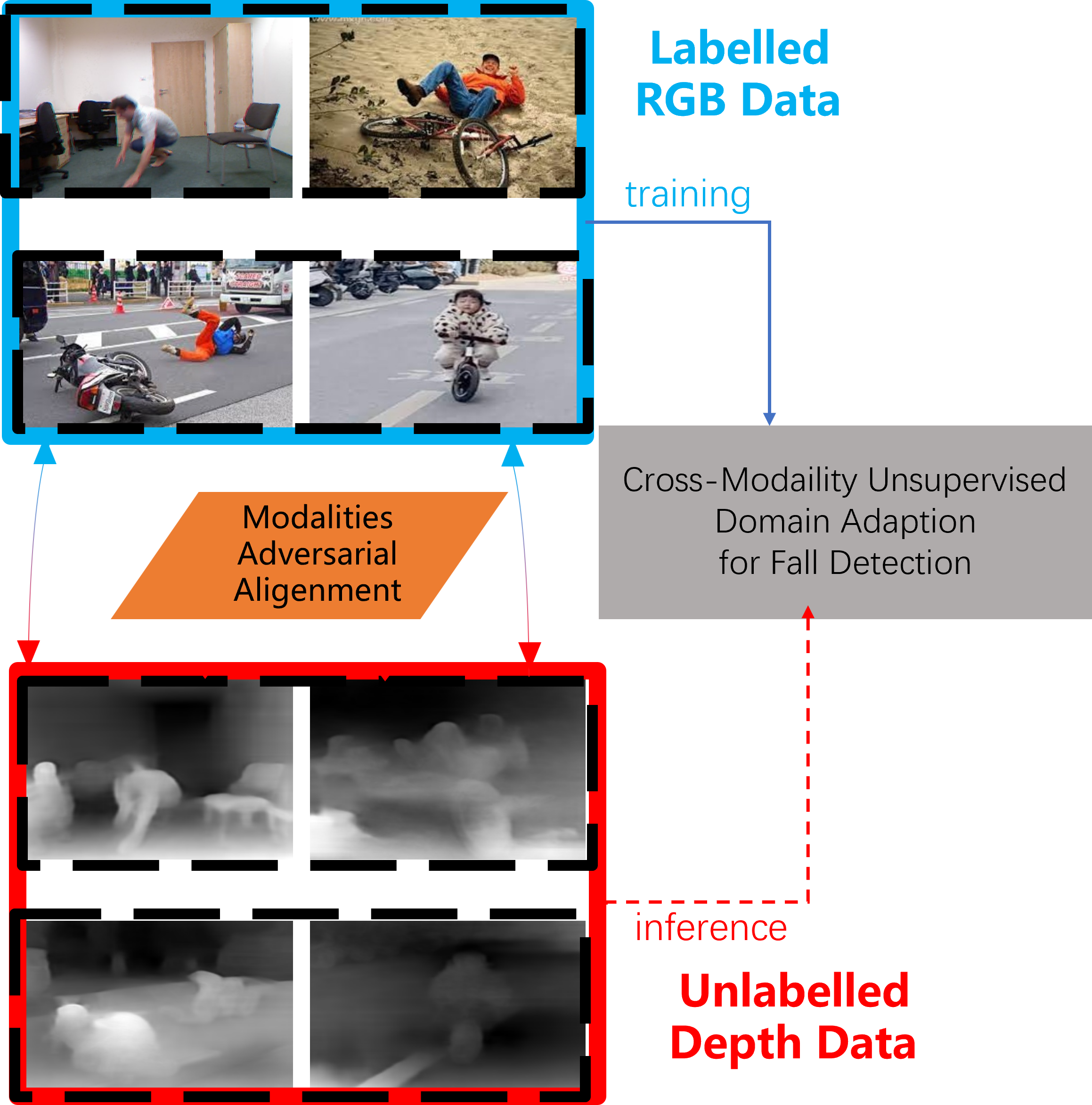}%
\end{wrapfigure}%
\begin{abstract}
Fall detection is a vital task in health monitoring, as it allows the system to trigger an alert and therefore enabling faster interventions when a person experiences a fall. 
Although most previous approaches rely on standard RGB video data, such detailed appearance-aware monitoring poses significant privacy concerns. 
Depth sensors, on the other hand, are better at preserving privacy as they merely capture the distance of objects from the sensor or camera, omitting color and texture information.

In this paper, we introduce a privacy-supporting solution that makes the RGB-trained model applicable in depth domain and utilizes depth data at test time for fall detection. 
To achieve cross-modal fall detection, we present an unsupervised RGB to Depth (RGB2Depth) cross-modal domain adaptation approach that leverages labelled RGB data and unlabelled depth data during training. Our proposed pipeline incorporates an intermediate domain module for feature bridging, modality adversarial loss for modality discrimination, classification loss for pseudo-labeled depth data and labeled source data, triplet loss that considers both source and target domains, and a novel adaptive loss weight adjustment method for improved coordination among various losses. Our approach achieves state-of-the-art results in the unsupervised RGB2Depth domain adaptation task for fall detection. Code is available at https://github.com/1015206533/privacy\_supporting\_fall\_detection.
\end{abstract}

\begin{IEEEkeywords}
Loss weight Adaptive, fall detection, unsupervised modality adaptation, intermediate domain module.
\end{IEEEkeywords}
\end{minipage}}}

\maketitle

\section{Introduction}
\label{sec:introduction}

According to the United Nations' predictions, 13\% of the global population was aged 60 or older~\cite{liu2014comparative}, and developing responsible technologies to support and assist the elderly becomes increasingly significant. Falls represent a major danger  for older adults, causing not only physical harm to the elderly but also to young individuals who live alone. As reported by the World Health Organization\footnote{https://www.who.int/news-room/fact-sheets/detail/falls}, falls lead to approximately 684,000 deaths annually and 37.3 million falls are severe enough to require medical attention for recovery.
There are different ways to approach fall detection, such as wearable equipment~\cite{1617246,yhdego2019towards, yacchirema2019fall, cao2016fall,aguiar2014accelerometer, lim2014fall, baek2013real}, using Wi-Fi  signal~\cite{chen2011wearable, wang2016wifall, ding2020wifi, wang2016rt,hu2020wifi} or  video monitoring systems~\cite{yajai2015fall, zhong2020multi, thummala2020fall, kottari2019real,cai2020vision, asif2020privacy, espinosa2019vision}. 
One advantage of video-based approaches is their physical practicality, as they do not impose physical burdens on users or require complex operational procedures compared to wearable devices~\cite{wang2016wifall}.
Video-based fall detection approaches typically build on well-established action recognition models to achieve accurate results~\cite{asif2020privacy}.

The majority of the current datasets and methodologies for fall detection rely on RGB data for training and evaluation~\cite{espinosa2019vision}. 
However, privacy protection has recently become an area of growing concern within the community, and the use of RGB-based data has been criticized for its potential to reveal detailed personal information.
As a consequence, there is a growing interest in privacy-preserving frameworks. 
In comparison to RGB data, depth data, also referred to as 3D data, is capable of representing the distance of objects from a camera or sensor. Depth data does not preserve detailed texture information, which significantly enhances its privacy-preserving characteristics. This alternative approach addresses privacy concerns while still providing valuable information for fall detection systems.
A fall detection method leveraging depth data at test-time would thereby be preferred according to the needs of users, given that it achieves accurate results. 
However, the existing depth-based datasets for fall detection are relatively small, providing limited training and test data for state-of-the art activity recognition architectures, which are known to be data-hungry. Considering the different privacy-preserving ability of different modality, people might choose to use different modality at the test time according to their needs. Cross-modal adaptation thereby becomes an important and challenging research direction of fall detection, which could make use of the well-established models pretrained on large-scale RGB-based datasets to realize depth-based fall detection at test time. 
In this work we focus on how to use labelled RGB data and unlabelled depth data for training and transfer knowledge from the RGB domain to the depth domain (RGB2Depth), which has been overlooked in the area of fall detection.

Since most of the depth-based fall detection dataset is small-scaled and can not be used as a sufficient test and training set while video-based approaches, \textit{e.g.}, X3D~\cite{feichtenhofer2020x3d}, always need large scale pretraining to achieve convergence, we first reformulate and adopt the existing activity recognition dataset, \textit{i.e.}, Kinetics~\cite{kay2017kinetics},for unsupervised domain adaptation (UDA) for the \textit{RGB2Depth fall detection} task. A subset of the data is transferred into Depth data through P\textsuperscript{2}Net~\cite{yu2020p} to provide sufficient test samples and well-established RGB-based pretrained weights are used to initialize the investigated models.
To bridge the gap between RGB domain and depth domain for fall detection, we establish the cross-modality unsupervised domain adaptation pipeline UMA-FD by using X3D~\cite{feichtenhofer2020x3d} model,  which stands as one of the most promising backbones for achieving accurate action recognition.
We first utilize an intermediate domain module~\cite{dai2021idm} to bridge representations of the  RGB and Depth domains, and then leverage multiple losses to constrain the latent space, \textit{e.g.,} an adversarial modality discrimination loss, triplet margin losses on the two domains, classification loss on the source RGB data and the pseudo-labeled depth data. 
Since different losses contribute in different ways, a fixed scheme to weight the losses may strongly restrict the attention of the learning process during different learning stages. Thereby, we propose an adaptive weighting approach for the employed loss functions. The network is asked to predict weighting parameters to adjust the weights of the losses using an additional multi-layer perceptron-based head. We have summarized our contributions as following,

\begin{itemize}
    \item We for the first time propose the RGB-to-Depth (RGB2Depth) unsupervised domain adaptation task in the context of fall detection and develop a new multi-source dataset and establishing a benchmark protocol for this purpose. 
    \item We also introduce a new pipeline to address this task. We employ 3D-CNN+LSTM~\cite{lu2018deep}, C3D~\cite{tran2015learning}, I3D~\cite{carreira2017quo}, X3D~\cite{feichtenhofer2020x3d} as our feature extraction backbones and utilize the intermediate domain module (IDM), modality adversarial alignment, and triplet margin loss to minimize the cross-domain gap. Additionally, we propose an adaptive weighting method for balancing the leveraged loss functions.
    \item Our model, UMA-FD, delivers the state-of-the art performance on the proposed RGB2Depth UDA task in comparison to other existing fall detection methods. Ablation studies showcase the efficiency  of the  proposed building block individually.
\end{itemize}

\section{Related Work}

\noindent\textbf{Fall detection.} The existing research  targeting fall detection can be clustered into three major groups based on the sensors employed. The first group focuses on wearable devices. The majority of these devices employ accelerometers as the primary sensor to capture signals from various body parts, such as the wrist, chest, and waist ~\cite{yhdego2019towards, yacchirema2019fall, cao2016fall,aguiar2014accelerometer, lim2014fall, baek2013real}. For example, Chen et al.~\cite{chen2019intelligent} used smartwatch on the wrist to monitor the movement status of indivudials. Mehmood et al.~\cite{mehmood2019novel} proposed a novel wearable sensor called SHIMMER to measure the signal from the waist.
The second group of fall detection research utilizes Wi-Fi signal networks ~\cite{chen2011wearable, wang2016wifall, ding2020wifi, wang2016rt,hu2020wifi, casilari2017umafall}. In~\cite{wang2016wifall}, Wang et al. proposed WiFall, a system that achieves fall detection by analyzing the correlation between radio signal variations and human activities.
Hu et al.~\cite{hu2020wifi} proposed DeFall, a system that operates based on an offline template-generating stage and an online decision-making stage, utilizing Wi-Fi features associated with human falls.
The third group involves vision-based approaches ~\cite{yajai2015fall, zhong2020multi, thummala2020fall, kottari2019real,cai2020vision, asif2020privacy, espinosa2019vision}, which typically use action recognition models as the feature extraction backbone for fall prediction.  Several existing datasets make video-based fall detection possible, e.g., UR Fall dataset~\cite{kwolek2014human}, Kinetics dataset~\cite{kay2017kinetics}, NTU dataset~\cite{liu2020ntu}, UP Fall dataset~\cite{martinez2019up}. 
For example, Khraief et al.~\cite{khraief2020elderly}  introduced a multi-stream deep convolutional neural network that employs both RGB and depth modalities for fall detection. Na et al.~\cite{lu2018deep} used 3D-CNN and LSTM as backbone to extract the features of RGB video and do fall detection. Chen et al.~\cite{chen2020vision} used an attention-guided bi-directional LSTM to achieve fall detection in complex backgrounds. \cite{asif2020privacy} takes privacy concerns into consideration and utilizes body skeletons and semantic segmentation masks as input to achieve fall detection, while discarding the numerous existing RGB data.
Compared to~\cite{asif2020privacy}, our RGB2Depth unsupervised domain adaptation approach aims to utilize as much existing data as possible for training the model. The model is expected to be adapted from the data-rich RGB domain to the privacy-preserving depth domain using unlabeled data. We do not leverage skeleton information due to the significant domain difference between RGB and skeleton data. Instead, we use depth data, as it provides a balance between preserving privacy and maintaining sufficient information for accurate domain mapping and learning from the data-rich RGB domain.

\noindent\textbf{Action recognition.}
Action recognition approaches can be categorized into video-based methods~\cite{arnab2021vivit, neimark2021video, bertasius2021space, li2022mvitv2, zhao2021tuber, fan2021multiscale, tong2022videomae, li2022mvitv2, liu2022swin, liu2022video, wang2022bevt} and skeleton-based methods ~\cite{shi2019two,liu2020disentangling,chen2021channel,shi2020skeleton,ye2020dynamic,zhang2020semantics,lee2022hierarchically}, where the latter group is  more closely related to our work. In video-based action recognition, the model is designed to learn underlying cues for human body actions from either individual still images or a sequence of frames. The existing well-established models include CNN-based models, \textit{e.g.,} C3D~\cite{tran2015learning}, I3D~\cite{carreira2017quo}, X3D~\cite{feichtenhofer2020x3d}, SlowFast~\cite{fan2020pyslowfast}, and transformer-based models, \textit{e.g.,}, MViTv2~\cite{li2022mvitv2} or Video Swin~\cite{liu2022swin}.

\noindent\textbf{Unsupervised domain adaptation.} The goal of unsupervised domain adaptation is to reduce the domain gap by utilizing unannotated data from the target domain. Due to its practicality in many applications, unsupervised domain adaptation has grasped the attention of  researchers from diverse fields, \textit{e.g.,} semantic segmentation and action recognition~\cite{song2021spatio, choi2020unsupervised, busto2018open, da2022dual,kim2021learning}. Unsupervised domain adaptation can be achieved, \textit{e.g.,}, via pseudo labelling~\cite{song2021spatio} and contrastive learning~\cite{kim2021learning, da2022dual,Schneider_2022_CVPR}. For cross modality domain adaptation, Ouyang et al.~\cite{ouyang2019data} proposed a variational encoder-based approach to adapt 3D MRI image into 3D CT image which is in medical imaging field. Dou proposed a plug-and-play domain adaptation module (DAM) targeting at cross-modality domain adaptation for biomedical segmentation. 
However, the research of  domain adaptation between different modalities in detecting falls has been very limited.
We thereby point it as an interesting research direction and present a framework for unsupervised domain adaptation for fall detection, which, to the best of our knowledge, is done for the first time.

\section{Dataset}\label{dataset_section}

In order to study cross-modality unsupervised domain adaptation from RGB2Depth for fall detection, we require multi-modal data that includes both RGB and Depth information in fall scenarios.
The Kinetics-700~ video dataset\cite{carreira2019short} contains 650,000 video clips, spanning 700 human action classes, with each clip annotated with an action class and lasting approximately 10 seconds.
From the Kinetics-700 dataset, we select two categories related to falling actions: the \textit{falling off bike} class and the \textit{falling off chair} class. The videos in these two categories serve as positive samples for fall detection. Additionally, we randomly select two categories, \textit{washing hands} and \textit{sweeping floor}, to be used as negative samples for fall detection.

\begin{figure}[t]
	\centering
    \hspace{0.01em}
	\begin{subfigure}{0.32\linewidth}
		\centering
		\includegraphics[width=0.95\linewidth]{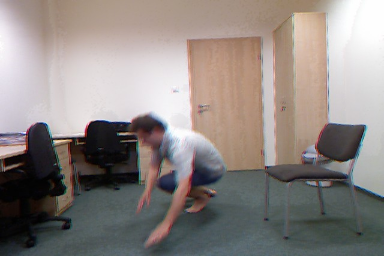}
		\label{rgbfig1}
	\end{subfigure}
	\begin{subfigure}{0.32\linewidth}
		\centering
		\includegraphics[width=0.95\linewidth]{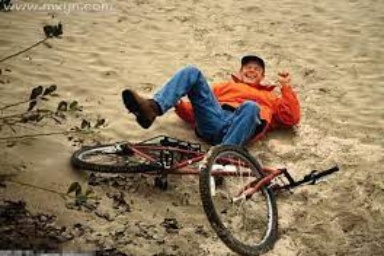}
		\label{rgbfig2}
	\end{subfigure}
    \begin{subfigure}{0.32\linewidth}
		\centering
		\includegraphics[width=0.95\linewidth]{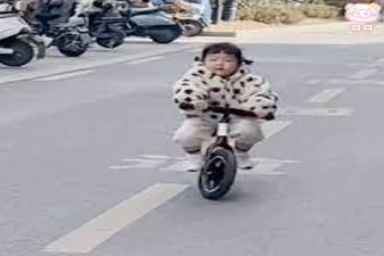}
		\label{rgbfig3}
	\end{subfigure}
    \vspace{0.02em}
    
    \hspace{0.01em}
	\begin{subfigure}{0.32\linewidth}
		\centering
		\includegraphics[width=0.95\linewidth]{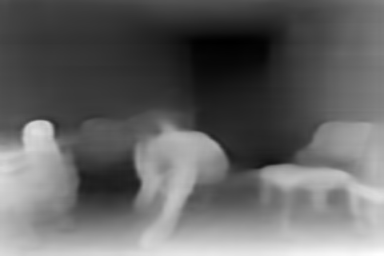}
		\label{depthfig1}
	\end{subfigure}
	\begin{subfigure}{0.32\linewidth}
		\centering
		\includegraphics[width=0.95\linewidth]{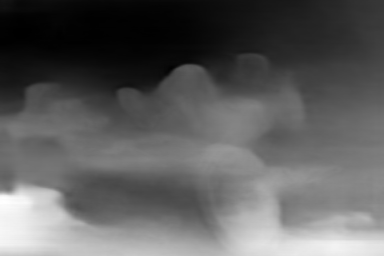}
		\label{depthfig2}
	\end{subfigure}
    \begin{subfigure}{0.32\linewidth}
		\centering
		\includegraphics[width=0.95\linewidth]{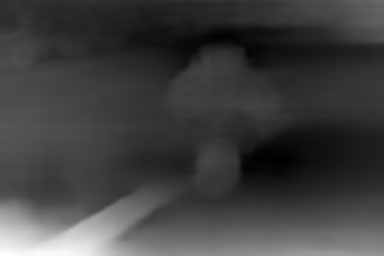}
		\label{depthfig3}
	\end{subfigure}
    \caption{The generated Depth data based on the P\textsuperscript{2}Net model. The first line is the original RGB frame images, and the second is the corresponding generated Depth data.}
	\label{rgb_depth_effect}
\vspace{-1em}
\end{figure}

\begin{table}[t]
\centering
\caption{Comparison with other fall detection dataset. For NTU RGB+D with multiple non-falling actions, we take the same number of other samples with falling action samples.}
\label{Comparison_dataset}
\setlength{\tabcolsep}{3pt}
\begin{tabular}{m{78pt}<{\centering} m{42pt}<{\centering} m{50pt}<{\centering} m{50pt}<{\centering}}
\hline
\vspace{0.5em}
~~
\vspace{0.5em}
&
Fall Samples
&
Other Samples
&
Total Number
\\
\hline
\vspace{0.5em}
UR Fall Detection~\cite{kwolek2014human}
\vspace{0.5em}
&
30
&
40
&
70 \\
\hline
\vspace{0.5em}
NTU RGB+D~\cite{shahroudy2016ntu}
\vspace{0.5em}
&
948
&
948
&
1896 \\
\hline
\vspace{0.5em}
Our Dataset
\vspace{0.5em}
&
1490&
1489&
2979 \\
\hline
\end{tabular}
\end{table}

\begin{table}[t]
\centering
\caption{The number of samples in our dataset.}
\label{table_dataset}
\setlength{\tabcolsep}{3pt}
\begin{tabular}{m{30pt}<{\centering} m{70pt}<{\centering} m{70pt}<{\centering} m{30pt}<{\centering}}
\hline
&
Training Set&
Test Set&
\vspace{0.2em}
Total Number \\
\hline
\vspace{0.3em}
Positive Sample
\vspace{0.2em}&
\vspace{0.3em}
falling off bike /678 falling off chair /612
\vspace{0.2em}&
\vspace{0.3em}
falling off bike /100 falling off chair /100
\vspace{0.2em}&
1490 \\
\hline
\vspace{0.3em}
Negative Sample
\vspace{0.2em}&
\vspace{0.3em}
washing hands /644 sweeping floor /645
\vspace{0.2em}&
\vspace{0.3em}
washing hands /100 sweeping floor /100
\vspace{0.2em}&
1489 \\
\hline
\vspace{0.3em}
Total Number&
2579&
400&
2979 \\
\hline
\end{tabular}
\end{table}

\begin{figure}[t]
\vspace{-2em}
\centerline{\includegraphics[width=0.8\columnwidth]{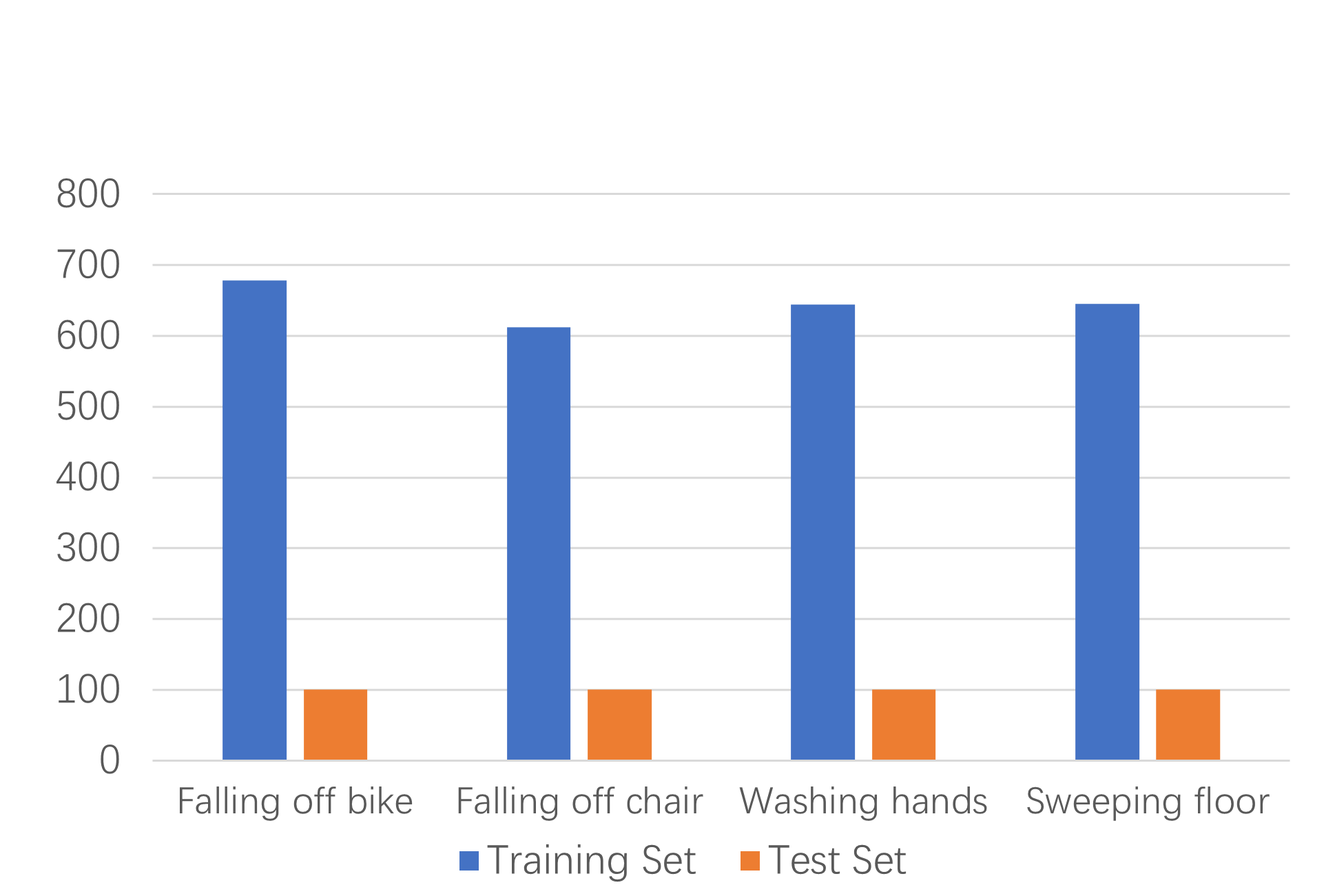}}
\caption{Data composition of training set and test set.}
\label{data_distribution}
\vspace{-1em}
\end{figure}

However, the four categories of videos mentioned above contain only RGB data. To generate corresponding Depth data, we can employ advanced depth estimation algorithms. We choose P\textsuperscript{2}Net~\cite{yu2020p} to generate corresponding 288x384 Depth data for each frame of each video in our dataset. Fig. ~\ref{rgb_depth_effect} shows three cases of the generated Depth data, demonstrating that the contour of the Depth data generally corresponds to the RGB data.
To standardize the data format, we resize each frame of the RGB video data to a size of 256x256. Ultimately, we generate a labeled fall detection dataset containing both RGB and Depth modalities.

Because the number of samples in the four categories of our dataset is unbalanced and we aim for  an equal number of positive and negative samples, we randomly sample the data of each category. 
We compare our dataset with two other existing datasets that include RGB videos and corresponding Depth data for fall detection in Table ~\ref{Comparison_dataset}. The UR Fall Detection dataset~\cite{kwolek2014human} contains only 30 falls and 40 activities of daily living sequences, which are insufficient to train a deep model.
The NTU RGB+D dataset~\cite{shahroudy2016ntu} is an action recognition dataset comprising 60 action classes, including 948 falling down action videos. We can select 948 videos of other actions as negative samples. However, these videos are limited to indoor scenes, resulting in a lack of variety. In contrast, our dataset contains 1490 fall samples and 1489 other samples, offering a larger dataset with richer and more diverse and unconstrained scenes compared to the other two datasets.

Next, we divide our dataset into training set and test set according to a certain ratio, and the number of training samples and test samples of each category after sampling of the dataset are provided in Table ~\ref{table_dataset} and Fig. \ref{data_distribution}.
We have a total of 2979 samples, including 1490 positive samples and 1489 negative samples.
We randomly select 100 samples from each category as the test set, resulting in 2579 training     and 400 test samples. Each sample contains RGB data and corresponding depth data.
The latter model evaluation experiments are all based on this dataset. Besides, we also use the NTU RGB+D dataset~\cite{shahroudy2016ntu} for latter experiments, which contains RGB videos and corresponding depth data, to make the experimental results more convincing.

\section{Proposed Method}
\label{sec:guidelines}
\begin{figure*}[!t]
\centerline{\includegraphics[width=18cm]{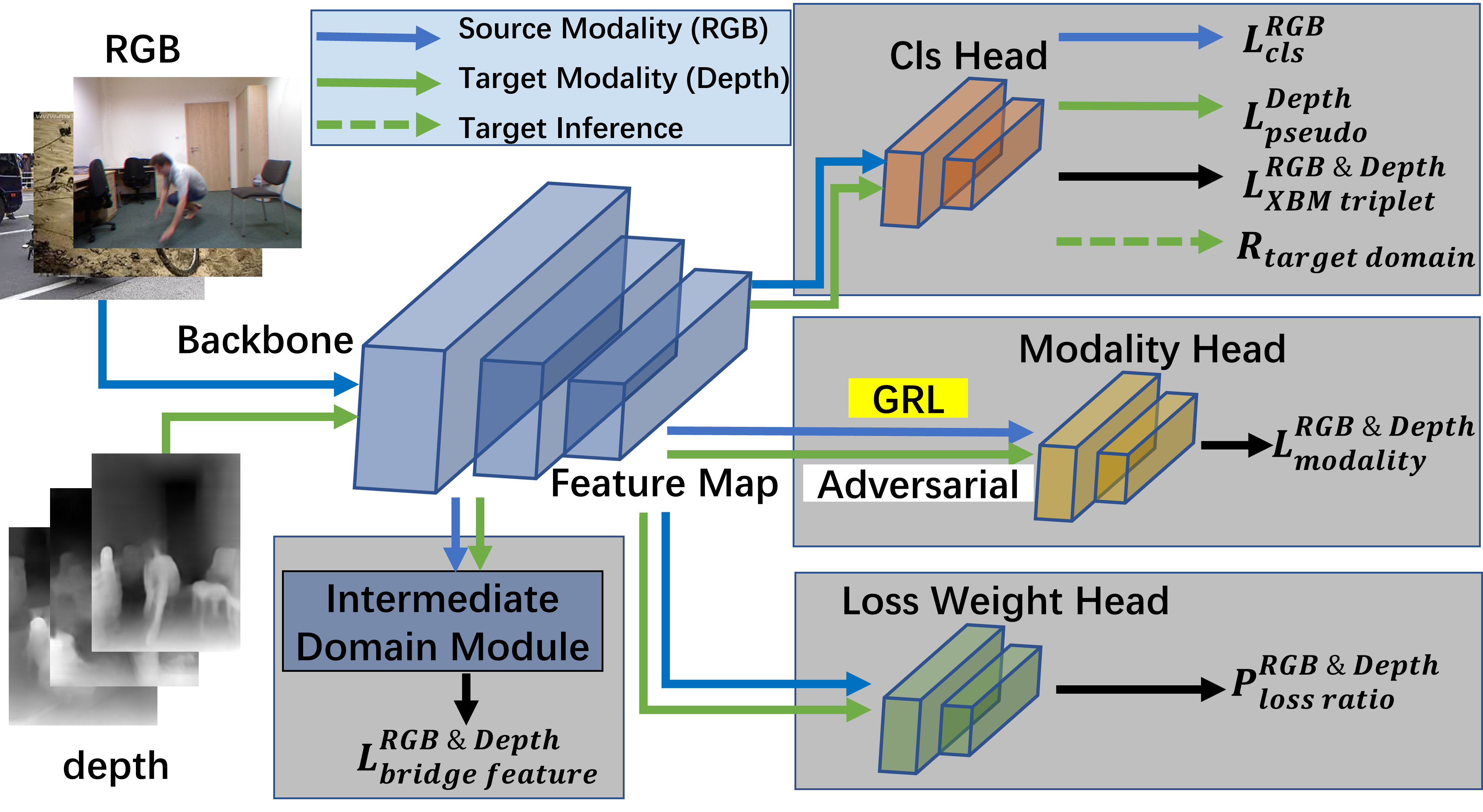}}
\caption{Proposed architecture: RGB data and depth data share the same backbone and pair randomly as input to generate the feature map of same format. IDM are incorporated in the backbone and generate bridge feature loss. Classification loss(Cls) Head, Modality Head and Loss Weight Head are the three head network to do label and modality classification and loss weight adaptation.}
\label{model_structure}
\end{figure*}

This section outlines our proposed method for RGB2Depth fall detection, which we refer to as \emph{Unsupervised Modality Adaptation for Fall Detection (UMA-FD)}.
In Fig. ~\ref{model_structure}, we present an overview of UMA-FD method, using cross-modality unsupervised domain adaptation to transfer the knowledge from the RGB source modality to the depth target modality.
In UMA-FD, we preprocess the data of the RGB and depth to generate a compatible input format and use a unified backbone network to generate the feature map of both data streams.
In the backbone, we incorporate intermediate domain module (IDM)\cite{dai2021idm} to generate the feature map of the intermediate modality, and then compute the bridge feature loss(IDM)\cite{dai2021idm}.
The classification layers of our network consists of thee heads: the label classification head, the modality head, and the loss weight adaptive head.
Next, we will first define the unsupervised modality adaptation problem, and then we  describe each building block of our proposed method in detail.

\subsection{Problem Definition}
Let $\textbf{X}$ and $\textbf{Y}$  represent input the data and the corresponding labels, respectively. Specifically, $X^R$ denotes RGB video data, while $X^D$ refers to depth sequence data. The labels $Y^R$ and $Y^D$ indicate whether the RGB and depth data represent a falling action, respectively.
Supervised learning aims to discover a representation, $G(\cdot)$, over learned features, $M(\cdot)$, that minimizes the empirical loss, $E_S[\mathcal{L}_y(G(M(x)), y)]$, given labeled samples ${(x, y)}$. This empirical loss is optimized over the labeled source modality, which in this case is the RGB data, $S=\{X^R, Y^R, \mathcal{B}^R\}$. Here, $\mathcal{B}^R$ denotes a distribution of RGB modality samples.
Modality adaptation seeks to minimize the empirical loss on the target modality, $T=\{X^D, Y^D, \mathcal{B}^D\}$, where $\mathcal{B}^D$ represents a distribution of depth modality samples. Notably, there is a significant difference between the source and target modality distributions, such that $\mathcal{B}^R \neq \mathcal{B}^D$.
In our proposed method for RGB2Depth fall detection task, the label space $Y^D$ is unknown and we use $S=\{X^R, Y^R, \mathcal{B}^R\}$ and $T=\{X^D, \mathcal{B}^D\}$ to train a model and predict $Y^D$.

\subsection{IDM and Bridge Feature Loss}
\begin{figure}[!t]
\centerline{\includegraphics[width=\columnwidth]{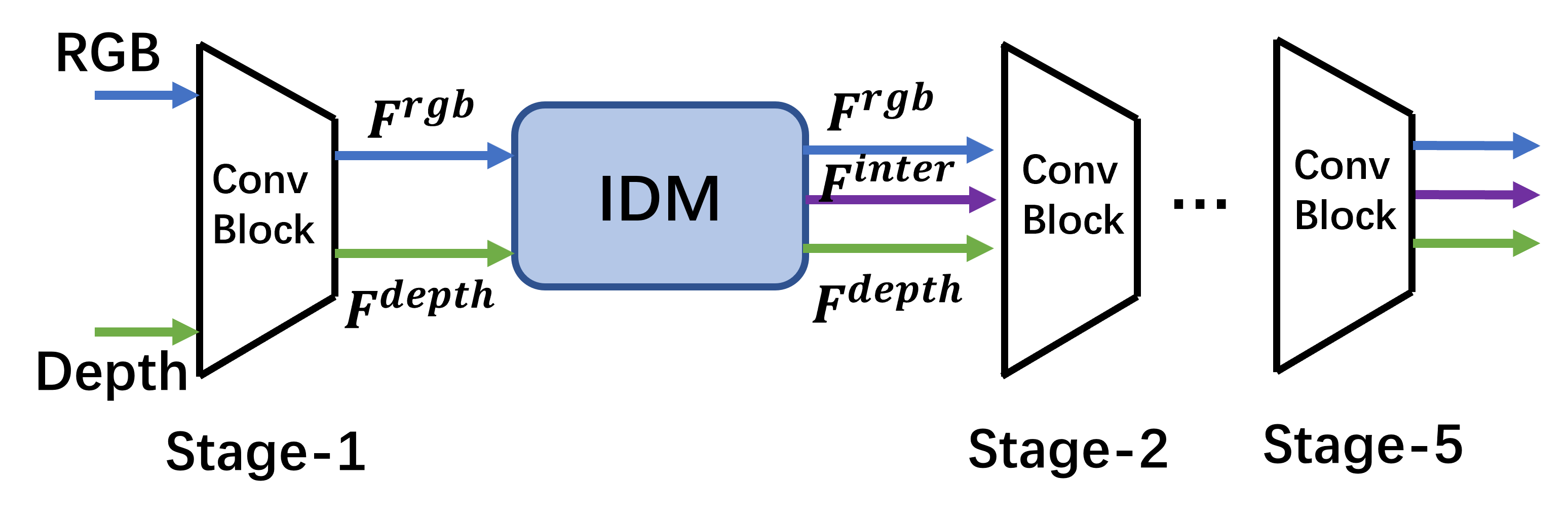}}
\caption{An overview of the location of IDM in the backbone.}
\label{IDM_module}
\end{figure}

During the training process, $X^R$ and $X^D$   are randomly paired as inputs to the backbone network.
The intermediate domain module (IDM) accepts data from both modalities as input and generates an intermediate modality feature map by weighted summation of the feature maps from the two modalities. The weighting coefficients are obtained through adaptive learning within the network.
IDM can be added between any two hidden layers of the backbone, generating hidden layer features for three modalities, including the intermediate modality, which are then fed into the subsequent backbone network layers. 
The intermediate modality feature map generated by IDM module can be represented as
\begin{equation}\begin{split}
\textbf{A}=\delta(MLP(FC([F_{h\_avg}^{R};F_{h\_max}^{R}])+\\FC([F_{h\_avg}^{D};F_{h\_max}^{D}]))).\label{eq1}\end{split}\end{equation}
\begin{equation}F^{inter}=\textbf{A}^{R} \cdot F_h^{R}+\textbf{A}^{D} \cdot F_h^{D}.\label{eq2}\end{equation}
where $F_{h}$ represents the feature map of of the hidden layer, subscript $avg$ and $max$ denote the average pooling and max pooling, $FC$ is a fully connected layer, $MLP$ stands for multiple fully connected layers, $[;]$ is the concatenation operator, and $\delta(\cdot)$ is the Softmax function.
In our case, different backbones are employed and we always add the IDM module after the first convolution block.
The placement of the IDM in backbone is illustrated in Fig. \ref{IDM_module}.

The backbone network generates the final feature maps for RGB, depth, and intermediate modalities. We employ the bridge feature loss~\cite{dai2021idm} to constrain the weighted sum of distances between the intermediate modality feature map and those of RGB and depth.
The weighted sum employs the weighting coefficients derived from the previously mentioned IDM module. This ensures that when the RGB modality has a significant influence on the intermediate modality, the bridge feature loss places greater emphasis on the distance between the RGB and intermediate modalities. Similarly, the same principle applies to the depth modality.
The bridge feature loss is computed as:
\begin{equation}\mathcal{L}_{bridge}^{R\&D}=\frac{1}{n}\sum_{i=1}^{n}\sum_{k\in\{R,D\}}a_i^k \cdot ||F^k-F^{inter}||_2.\label{eq3}\end{equation}
where $F$ is the feature map of the final output of backbone, $a$ is the weighting coefficient generated by the IDM module, and $||\cdot||$ represents the 2-norm to calculate the spatial Euclidean distance between two feature maps.

The bridge feature loss  ensures that the feature map of the intermediate modality is in the line between the RGB modality and the depth modality in the spatial distribution, and then constrains the backbone to learn the appropriate feature map of the RGB modality and the depth modality.

\subsection{Unsupervised Modality Adaptation}
Even through the source modality and the target modality are different, the underlying clues which are used for fall detection have strong potential for correlations. In this case, supervised training on the source modality can help to delve the informative cues on the target modality. Motivated by this, our method also  minimises both the classification loss of source modality and the distribution discrepancy between the source and target modality. 

Following the backbone,  a supervised classification head is constructed using a fully connected network.
 For the RGB modality data, which has associated sample labels, we can compute the cross-entropy loss:
\begin{equation}\mathcal{L}_{cls}^{R}=\sum-ylogQ(X^{R}).\label{eq31}\end{equation}
\begin{equation}Q(x)=\sigma(G(M(x))).\label{eq5}\end{equation}
where $G(\cdot)$ is the classification head, $M(\cdot)$ is the backbone, and $\sigma(\cdot)$ is the sigmoid function.
Since there are no labels available for the depth  data, a threshold-based pseudo-labeling technique is employed for supervision, allowing us to obtain the classification loss on the depth data.
With this approach, we estimate  pseudo-labels for the depth samples, which meet the threshold condition:

\begin{equation}
Y_{pseudo}^D=
\left\{
             \begin{array}{lr}
             0, & \sigma(G(M(X^D))) \leq 1-\tau \\
             & \\
             1, & \hspace{-1em} \sigma(G(M(X^D))) \geq \tau \\
             \end{array}
\right.
\end{equation}
where $\tau$ is the pseudo label threshold.
We then calculate the corresponding pseudo-label cross-entropy loss
\begin{equation}\mathcal{L}_{pseudo}^{D}=\sum_{X^{D} \in T^{part}}-y^{pseudo}logQ(X^{D}) .\label{eq6}\end{equation}
where $T^{part}$ represents the depth samples set that meets the threshold condition.

Triplet loss~\cite{hermans2017defense} is a commonly used loss for metric learning, and can be also used to deal with classification problem.
Due to the use of a 3D convolutional network model for video data and the limited memory of a single GPU, the batch size of training data is quite small. This makes it challenging to compute the triplet loss within a single batch. The cross-batch memory mechanism (XBM)~\cite{wang2020cross} addresses this issue by serving as a module that stores previously processed training data during the training process.
During the training process, the XBM component retains the feature maps from previous batches. Triplet loss is then calculated based on the current batch's feature maps and those stored in the XBM. As a result, we obtain the XBM\_triplet loss, which can be represented as:
\begin{equation}\begin{split}\mathcal{L}_{XBM\_triplet}^{R\&D}=max(d(F_{cur}, F_{pre}^p) - d(F_{cur}, F_{pre}^n) \\ + margin, 0) .\label{eq7}\end{split}\end{equation}
where $F_{cur}$ represents the feature map of the current sample, and $F_{pre}^p, F_{pre}^n$ respectively represent the previous sample features with the same label and different label with the current sample respectively.
The XBM\_triplet loss constraints the maximum distance of feature map between samples with same label smaller than the minimum distance between samples with different label, which improves the discriminative ability of the learned latent representations.

\subsection{Modalities Adversarial Alignment}
In unsupervised domain adaptation, both generative and discriminative adversarial approaches have been proposed for bridging the distribution discrepancy between source and target domains.
For high-dimensional data streams, such as video, discriminative approaches are more suitable~\cite{munro2020multi}.
Discriminative methods train a discriminator, $C(\cdot)$, to predict the modality of an input from the learnt features, $M(\cdot)$.
By maximising the discriminator loss, the network learns a feature representation that is invariant to both modalities.

In our scenario, to align the RGB and depth data, we propose a modality discriminator that penalizes feature variability between the modalities. 
The modality discriminator, $C(\cdot)$, contains a Gradient Reversal Layer(GRL)~\cite{li2018unsupervised} and a fully connected network to learn the modality  representation $M(\cdot)$.
Given a binary modality label, $y_d$, indicating if a sample $X$ belongs to the RGB or depth domain, we propose the following  modality loss:
\begin{equation}\begin{split}\mathcal{L}_{modality}^{R\&D}=\sum_{k \in \{R, D\}} -y_dlog(C(M(X^k)))- \\ (1-y_d)log(1-C(M(X^k))) .\label{eq8}\end{split}\end{equation}

The $\mathcal{L}_{modality}^{R\&D}$ loss reduces the variance between different modalities in the backbone's feature maps. This ensures that the features trained on the RGB data become more applicable to the depth data.

\subsection{Total Loss and Loss Weight Adaptation}
To summarize the above components, the final loss can be expressed as follows:
\begin{equation}\begin{split}\mathcal{L}=\lambda_a \mathcal{L}_{cls}^{R} + \lambda_b \mathcal{L}_{pseudo}^{D} + \lambda_c \mathcal{L}_{modality}^{R\&D} + \\ \lambda_d \mathcal{L}_{bridge}^{R\&D} + \lambda_e \mathcal{L}_{XBM\_triplet}^{R\&D} .\label{eq9}\end{split}\end{equation}
where $\lambda_a, \lambda_b, \lambda_c, \lambda_d, \lambda_e$ are the proportional coefficients.
During the training process, the overall network consists of five losses, and the impact of each loss on the final depth data's classification accuracy is unknown. We therefore need to adjust the value of $\lambda_a, \lambda_b, \lambda_c, \lambda_d, \lambda_e$ accordingly.
Manual adjustment can be time-consuming and may not yield an optimal combination solution.
To address this  problem, we consider using an adaptive network to automatically learn the loss weight, aiming to obtain the optimal solution. 
The loss weight adaptive network $W(\cdot)$ consists of a three-layer fully connected network and the corresponding activation functions. The network output is a five-dimensional weight coefficient
\begin{equation}P=softmax(W(M(X))) .\label{eq10}\end{equation}
Then, the final loss of the adaptive network becomes:
\begin{equation}\begin{split}\mathcal{L}=P_1 \cdot \mathcal{L}_{cls}^{R} + P_2 \cdot \mathcal{L}_{pseudo}^{D} + P_3 \cdot \mathcal{L}_{modality}^{R\&D} + \\ P_4 \cdot \mathcal{L}_{bridge}^{R\&D} + P_5 \cdot \mathcal{L}_{XBM\_triplet}^{R\&D} .\label{eq11}\end{split}\end{equation}

\section{Experiments and Results}
In this section, we first discuss the implementation details and evaluation metrics. Then, we evaluate our proposed method UMA-FD and compare the results with the  baseline  and the supervised target method for fall detection on the NTU RGB+D dataset~\cite{shahroudy2016ntu} and our dataset. In order to make the results more convincing, we compare the outcome of a fall detection backbone 3D-CNN+LSTM~\cite{lu2018deep} and other three different CNN-based backbones including C3D~\cite{tran2015learning}, I3D~\cite{carreira2017quo}, X3D~\cite{feichtenhofer2020x3d}.
Next, various ablation experiments are implemented and their results are discussed. Finally, we showcase  qualitative results of UMA-FD and analyze the classification results of some samples.

\subsection{Implementation Details and Evaluation Metrics}
During training process, we use the dataset described in Section \ref{dataset_section}. The NTU RGB+D dataset contains 948 falling down videos and corresponding depth data as positive samples, 948 other action videos and corresponding depth data as negative samples. We choose 180 positive samples and 180 negative samples as test set and all other samples as training set. Our generated dataset consists of RGB video data of four categories from the Kinetics-700 database along with  the depth data produced for each frame of video data with a  depth estimation model.
Our training set contains $2579\times2=5158$ samples  from RGB and depth modalities in total, of which 2579 RGB examples are labelled (source modality), and the 2579 depth modality clips are unlabelled (target modality).
The test set has 400 depth modality samples in total and is generated by randomly sampling 100 clips from each of the four categories including 200 positive examples (fall action) and 200 negative examples (other action).

All experiments are conducted using two NVIDIA GeForce RTX 3090 GPUs with 24G memory for parallel training and testing.
During the training process,  RGB and depth data are randomly paired, and each batch contains an RGB sample and a depth sample. For training, we use the open source framework mmaction2~\cite{2020mmaction2}, which contains  pre-trained models  C3D~\cite{tran2015learning}, I3D~\cite{carreira2017quo} and X3D~\cite{feichtenhofer2020x3d}.
When training  for the first time, we initialize the weights using pre-trained models provided in~\cite{2020mmaction2}, while each  network head is randomly initialized using a normal distribution with a mean of 0 and a standard deviation of 0.01. For each ablation experiment in Table ~\ref{table_ablation}, the initialization of the model parameters is based on the best parameter configuration obtained from the past experiments.
The optimizer for all experiments is SGD~\cite{amari1993backpropagation} with momentum of 0.9. 
For I3D and X3D backbones, the initial learning rate is 0.0001. For C3D, the initial learning rate is set  to 0.001. We train our model for 120 epochs for each group of experiments. After training for 60 epochs, the learning rate decays to one-tenth of the original value. For the pseudo-label threshold, we use 0.8, 0.8, 0.8, and 0.7 respectively for the fall detection backbone method 3D-CNN+LSTM and three different backbones C3D, I3D, and X3D.

Based on the dataset labels, the fall sample is defined as a positive sample, and the non-fall sample is used as a negative sample, and the number of positive and negative samples is equal. For binary classification problems, commonly employed evaluation metrics include classification accuracy, positive sample precision rate, positive sample recall rate, F1 score (which takes into account both positive sample precision and recall rates), and AUC (which evaluates the ranking of positive and negative sample scores).
Except for AUC, the other metrics depend on the threshold used for distinguishing between positive and negative sample scores. During the evaluation process, we set the score threshold between positive and negative samples at 0.5. We then calculate each evaluation metric based on the positive and negative sample labels predicted by this threshold.

\subsection{Compared with Baseline and Supervised Target Method }
Since fall detection through cross-modality unsupervised adaptive learning is a novel problem, there are no existing related works that can serve as a baseline for our research. Therefore, we need to define a baseline based on our dataset.
The most straightforward approach, utilizing the concept of transfer learning, involves training a fall detection model on the labeled RGB data and then directly predicting using the unlabeled depth dataset. We define this as the baseline method.
Additionally, we can obtain results from a supervised target method, in which we assume the depth data labels are known. In this supervised target method, we use both RGB and depth data, which are labeled, to learn a binary classification model. We then predict the depth data to obtain the classification results. To make the results more convincing, we conduct comparative experiments based on a fall detection backbone 3D-CNN+LSTM~\cite{lu2018deep} and other three different CNN-based backbones including C3D~\cite{tran2015learning}, I3D~\cite{carreira2017quo}, X3D~\cite{feichtenhofer2020x3d}.

The results comparison based on the NTU RGB+D dataset and our generated kinetics dataset are respectively provided in Table \ref{table_results_comparison_NTUD} and Table \ref{table_results_comparison}.
{On the NTU RGB+D dataset, compared with the baselines under 3D-CNN+LSTM and C3D, I3D, and X3D backbones on the constructed fall detecton benchmark, the accuracy of UMA-FD is increased by 10.83\%, 5.95\%, 6.66\%, 10.83\%, the F1 scores are increased by 45.05\%, 9.06\%, 9.89\% and 10.61\%, and the AUC is increased by 5.96\%, 20.16\%, 20.77\% and 6.63\%, respectively. Note that, 3D-CNN+LSTM is an well established fall detection backbone proposed by~\cite{lu2018deep} while the others are backbones designed for general action recognition.}
{On our generated kinetics dataset, compared with the baseline using the fall detection backbone, \textit{i.e.}, 3D-CNN+LSTM,  and conventional activity recognition backbones, \textit{i.e.}, C3D, I3D, and X3D , the accuracy of UMA-FD is increased by 3.25\%, 4.75\%, 4.25\%, 5.5\%, the F1 scores are increased by 10.56\%, 5.14\%, 9.19\% and 8.61\%, and the AUC is increased by 4.19\%, 3.92\%, 3.12\% and 4.26\%, respectively.} The results consistently showcase that through our  cross-modality unsupervised adaptation  methods, the fall detection performance on depth data is significantly improved. Furthermore, this improvement is independent of the specific backbone used, which demonstrates  versatility of  our  method.
The outcome of different backbones are also  in line with expectations.
The I3D backbone has fewer parameters and achieves nearly equivalent results compared to the C3D backbone. Meanwhile, the X3D backbone is one of the most promising existing backbones for action recognition, yielding significantly better results than the other two backbones.
On the other hand, compared with supervised target method, since there is no label information of depth data, the accuracy of UMA-FD is still much lower.
There is still room for improvement regarding the accuracy of our method, which will also be the focus of our follow-up research. 
Since the X3D backbone yields the best results, all subsequent experiments will be based on the X3D backbone.

\begin{table}
\centering
\caption{Results comparison of the NTU RGB+D dataset under a well-established fall detection backbone, \textit{i.e.},  3D-CNN+LSTM, and other different backbones including C3D, I3D and X3D. Under each set of experiments, we compare the results of our proposed method UMA-FD with baseline and supervised target method.}
\label{table_results_comparison_NTUD}
\begin{tblr}{
  cell{1}{1} = {r=2}{},
  cell{1}{2} = {c=5}{},
  cell{6}{1} = {r=2}{},
  cell{6}{2} = {c=5}{},
  cell{11}{1} = {r=2}{},
  cell{11}{2} = {c=5}{},
  cell{16}{1} = {r=2}{},
  cell{16}{2} = {c=5}{},
  row{4,9,14,19} = {bg = lightgray},
  vline{2} = {1-20}{},
  hline{1,3-6,8-11,13-16,18-21} = {-}{},
  hline{2,7,12,17} = {2-6}{},
}
~ ~ methods         & ~ ~ ~ ~ ~ ~ ~ ~ ~ ~ ~\textbf{3D-CNN+LSTM~\cite{lu2018deep}} &                 &                 &                 &                 \\
                    & Accuracy                                  & Precision       & Recall          & F1 Score        & AUC             \\
~ ~ Baseline        & ~~57.78                                    & ~~91.18          & 17.22          & ~~28.97          & 74.49          \\
\textbf{~ ~ UMA-FD} & ~~\textbf{68.61}                           & ~~\textbf{63.14} & \textbf{89.44} & ~~\textbf{74.02} & \textbf{80.45} \\
Supervised Target   & ~~94.44                                    & ~~95.45          & 93.33          & ~~94.38          & 98.55          \\
~ ~ methods         & ~ ~ ~ ~ ~ ~ ~ ~ ~ ~ ~ ~ ~ ~ ~\textbf{C3D~\cite{tran2015learning}} &                 &                 &                 &                 \\
                    & Accuracy                                  & Precision       & Recall          & F1 Score        & AUC             \\
~ ~ Baseline        & ~~60.32                                    & ~~58.36          & 73.42          & ~~65.03          & 62.90          \\
\textbf{~ ~ UMA-FD} & ~~\textbf{66.27}                           & ~~\textbf{61.93} & \textbf{92.18} & ~~\textbf{74.09} & \textbf{83.06} \\
Supervised Target   & ~~95.96                                    & ~~98.48          & 94.27          & ~~96.33          & 98.37          \\
~ ~ methods         & ~ ~ ~ ~ ~ ~ ~ ~ ~ ~ ~ ~ ~ ~ ~\textbf{I3D~\cite{carreira2017quo}} &                 &                 &                 &                 \\
                    & Accuracy                                  & Precision       & Recall          & F1 Score        & AUC             \\
~ ~ Baseline        & ~~59.17                                    & ~~57.27          & 72.22          & 63.88          & 61.20          \\
\textbf{~ ~ UMA-FD} & ~~\textbf{65.83}                           & ~~\textbf{59.86} & \textbf{96.11} & ~~\textbf{73.77} & \textbf{81.97} \\
Supervised Target   & ~~95.83                                    & ~~98.25          & 93.33          & ~~95.73          & 98.16          \\
~ ~ methods         & ~ ~ ~ ~ ~ ~ ~ ~ ~ ~ ~ ~ ~ ~ ~\textbf{X3D~\cite{feichtenhofer2020x3d}} &                 &                 &                 &                 \\
                    & Accuracy                                  & Precision       & Recall          & F1 Score        & AUC             \\
~ ~ Baseline        & ~~83.89                                    & ~~82.80          & 85.56          & ~~84.16          & 91.63          \\
\textbf{~ ~ UMA-FD} & ~~\textbf{94.72}                           & ~~\textbf{93.99} & \textbf{95.56} & ~~\textbf{94.77} & \textbf{98.26} \\
Supervised Target   & ~~98.33                                    & ~~98.33          & 98.33          & ~~98.33          & 99.50          
\end{tblr}
\end{table}

\begin{table}
\centering
\caption{Results comparison of our generated kinetics dataset under a well-established fall detection backbone, \textit{i.e.},  3D-CNN+LSTM, and other different backbones including C3D, I3D and X3D. Under each set of experiments, we compare the results of our proposed method UMA-FD with baseline and supervised target method.}
\label{table_results_comparison}
\begin{tblr}{
  cell{1}{1} = {r=2}{},
  cell{1}{2} = {c=5}{},
  cell{6}{1} = {r=2}{},
  cell{6}{2} = {c=5}{},
  cell{11}{1} = {r=2}{},
  cell{11}{2} = {c=5}{},
  cell{16}{1} = {r=2}{},
  cell{16}{2} = {c=5}{},
  row{4,9,14,19} = {bg = lightgray},
  vline{2} = {1-20}{},
  hline{1,3-6,8-11,13-16,18-21} = {-}{},
  hline{2,7,12,17} = {2-6}{},
}
~ ~ methods         & ~ ~ ~ ~ ~ ~ ~ ~ ~ ~ ~\textbf{3D-CNN+LSTM~\cite{lu2018deep}} &                 &                 &                 &                 \\
                    & Accuracy                                  & Precision       & Recall          & F1 Score        & AUC             \\
~ ~ Baseline        & ~~61.25                                    & ~~67.79          & 50.50          & ~~57.88          & 65.74          \\
\textbf{~ ~ UMA-FD} & ~~\textbf{64.50}                           & ~~\textbf{61.60} & \textbf{77.00} & ~~\textbf{68.44} & \textbf{69.93} \\
Supervised Target   & ~~73.25                                    & ~~75.14          & 69.50          & ~~72.21          & 81.18          \\
~ ~ methods         & ~ ~ ~ ~ ~ ~ ~ ~ ~ ~ ~ ~ ~ ~ ~\textbf{C3D~\cite{tran2015learning}} &                 &                 &                 &                 \\
                    & Accuracy                                  & Precision       & Recall          & F1 Score        & AUC             \\
~ ~ Baseline        & ~~67.00                                    & ~~69.32          & 61.00          & ~~64.89          & 72.67          \\
\textbf{~ ~ UMA-FD} & ~~\textbf{71.75}                           & ~~\textbf{74.58} & \textbf{66.00} & ~~\textbf{70.03} & \textbf{76.59} \\
Supervised Target   & ~~79.75                                    & ~~78.47          & 82.00          & ~~80.20          & 85.43          \\
~ ~ methods         & ~ ~ ~ ~ ~ ~ ~ ~ ~ ~ ~ ~ ~ ~ ~\textbf{I3D~\cite{carreira2017quo}} &                 &                 &                 &                 \\
                    & Accuracy                                  & Precision       & Recall          & F1 Score        & AUC             \\
~ ~ Baseline        & ~~68.00                                    & ~~76.09          & 52.50          & ~~62.13          & 74.70          \\
\textbf{~ ~ UMA-FD} & ~~\textbf{72.25}                           & ~~\textbf{73.80} & \textbf{69.00} & ~~\textbf{71.32} & \textbf{77.82} \\
Supervised Target   & ~~79.25                                    & ~~80.95          & 76.50          & ~~78.66          & 86.33          \\
~ ~ methods         & ~ ~ ~ ~ ~ ~ ~ ~ ~ ~ ~ ~ ~ ~ ~\textbf{X3D~\cite{feichtenhofer2020x3d}}  &                 &                 &                 &                 \\
                    & Accuracy                                  & Precision       & Recall          & F1 Score        & AUC             \\
~ ~ Baseline        & ~~73.00                                    & ~~78.75          & 63.00          & ~~70.00          & 81.79          \\
\textbf{~ ~ UMA-FD} & ~~\textbf{78.50}                           & ~~\textbf{78.22} & \textbf{79.00} & ~~\textbf{78.61} & \textbf{86.05} \\
Supervised Target   & ~~92.50                                    & ~~93.37          & 91.50          & ~~92.43          & 97.90
\end{tblr}
\end{table}
\begin{table}[t]
\caption{Ablation of our proposed method UMA-FD, showing the contribution of the various module and corresponding loss functions.}
\label{table_ablation}
\setlength{\tabcolsep}{3pt}
\begin{tabular}{m{55pt}<{\centering} m{33pt}<{\centering} m{33pt}<{\centering} m{33pt}<{\centering} m{33pt}<{\centering} m{33pt}<{\centering}}
\hline
\vspace{0.3em}
~~
\vspace{0.2em}
&
Accuracy&
Precision&
Recall&
F1 Score&
AUC \\
\hline
\vspace{0.3em}
Baseline (V-01)
\vspace{0.2em}&
73.00&
78.75&
63.00&
70.00&
81.79 \\
\hline
+Modality Loss (V-02)&
75.25&
77.90&
70.50&
74.02&
83.04 \\
\hline
+Pseudo Loss (V-03)&
76.25&
72.34&
\textbf{85.00}&
78.16&
83.05 \\
\hline
+Bridge Feature Loss (V-04)&
77.25&
\textbf{82.25}&
69.50&
75.34&
84.67 \\
\hline
+XBM\_Triplet Loss (V-05)&
77.75&
79.68&
74.50&
77.00&
84.10 \\
\hline
\rowcolor[HTML]{C0C0C0}
\vspace{0.6em}
{\normalsize UMA-FD (V-06)}
\vspace{0.3em}&
\vspace{0.2em}
{\normalsize \textbf{78.50}}&
\vspace{0.2em}
{\normalsize 78.22}&
\vspace{0.2em}
{\normalsize 79.00}&
\vspace{0.2em}
{\normalsize \textbf{78.61}}&
\vspace{0.2em}
{\normalsize \textbf{86.05}}
\\
\hline
\end{tabular}
\label{tab1}
\end{table}
\begin{figure}[t]
\centerline{\includegraphics[width=\columnwidth]{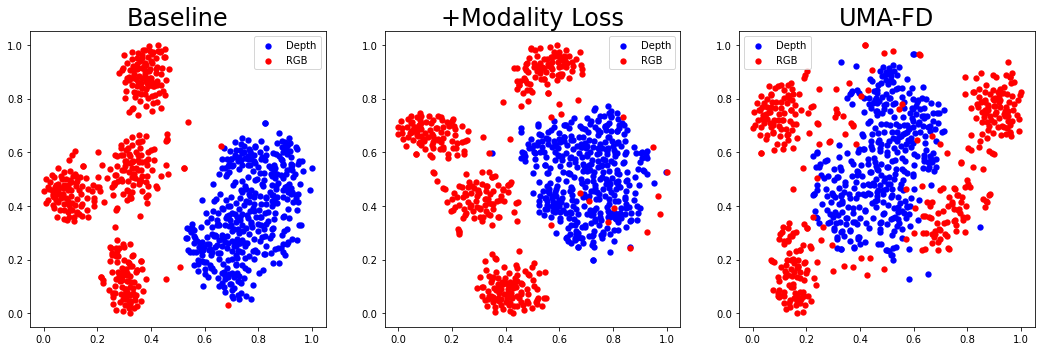}}
\caption{t-SNE plots of RGB data (red points) and Depth data (blue point) feature spaces produced by baseline, the method of adding modality loss and our proposed method UMA-FD.}
\label{tSNE_analysis}
\end{figure}
\begin{figure*}[t]
    \centering
    \includegraphics[width=1.0\linewidth]{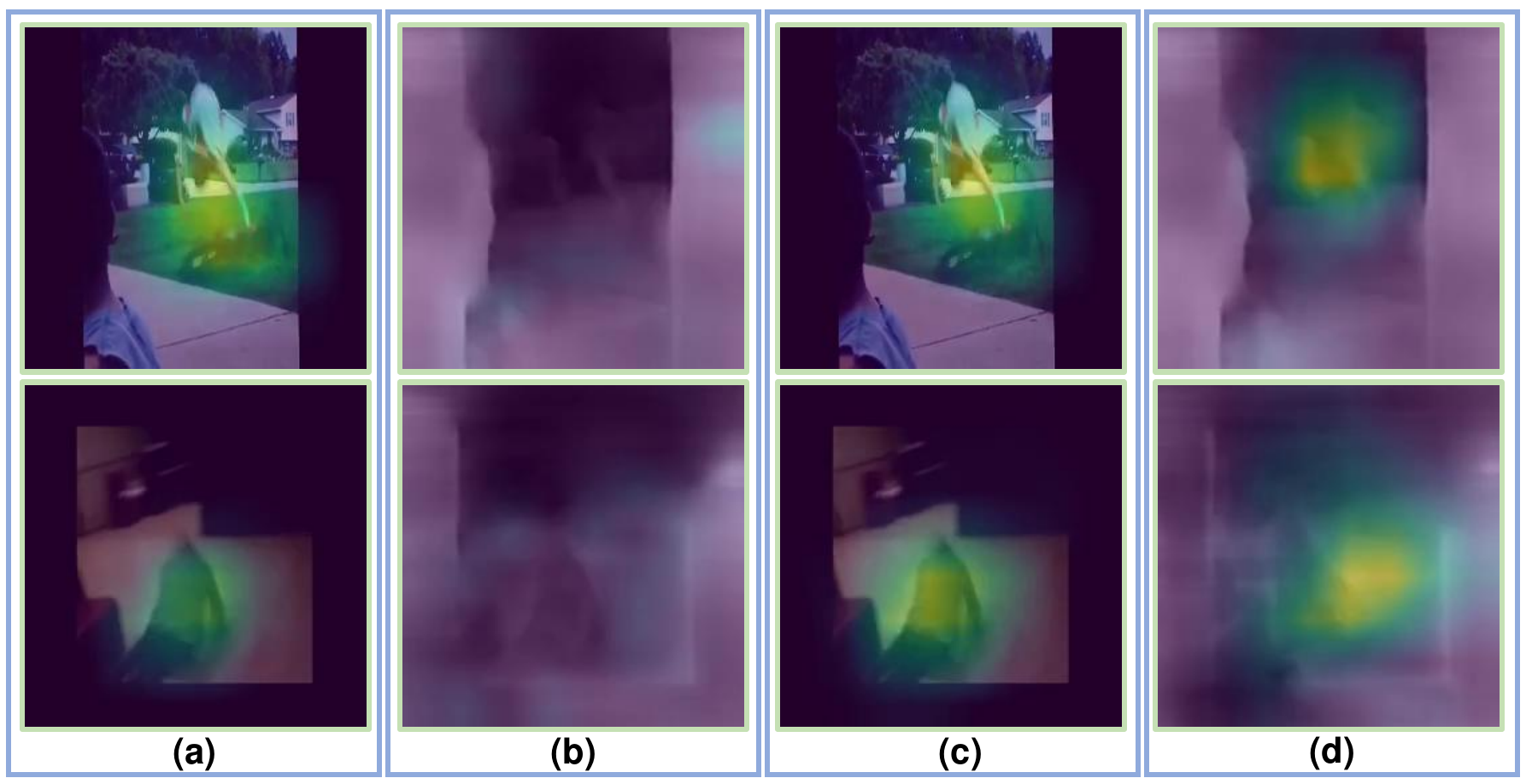}
    \caption{An overview of the model interpretability.
    GradCAM-based visualization resutls shown for RGB baseline (a), Depth baseline (b), for the RGB-based UMA-FD (c) and the Depth-based UMA-FD (d). 
    }
\label{model_interpretability}
\end{figure*}
\begin{figure*}[t]
    \centering
    \includegraphics[width=1\linewidth]{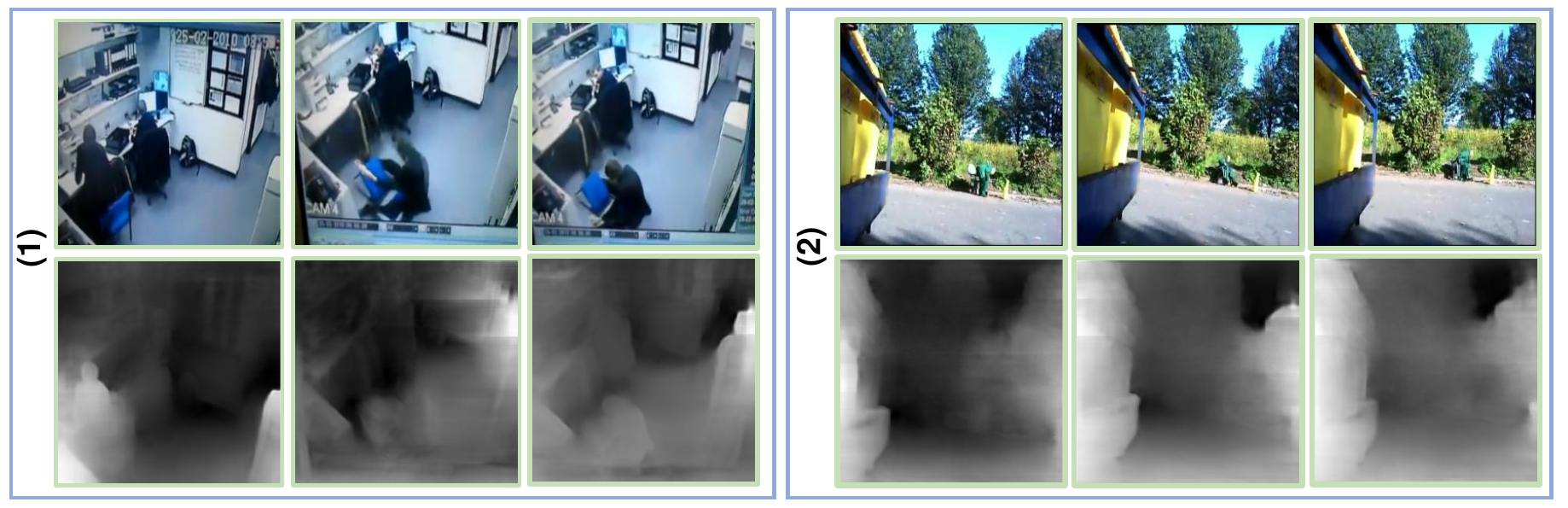}
    \caption{Two incorrect classification cases: Here are two examples where the classification results of UMA-FD are incorrect. Each example contains three frames extracted from the video. The first row represents RGB data, while the second row displays the corresponding depth data.}
    \label{single_case_error}
\end{figure*}
\begin{figure*}[t]
    \centering
    \includegraphics[width=1.0\linewidth]{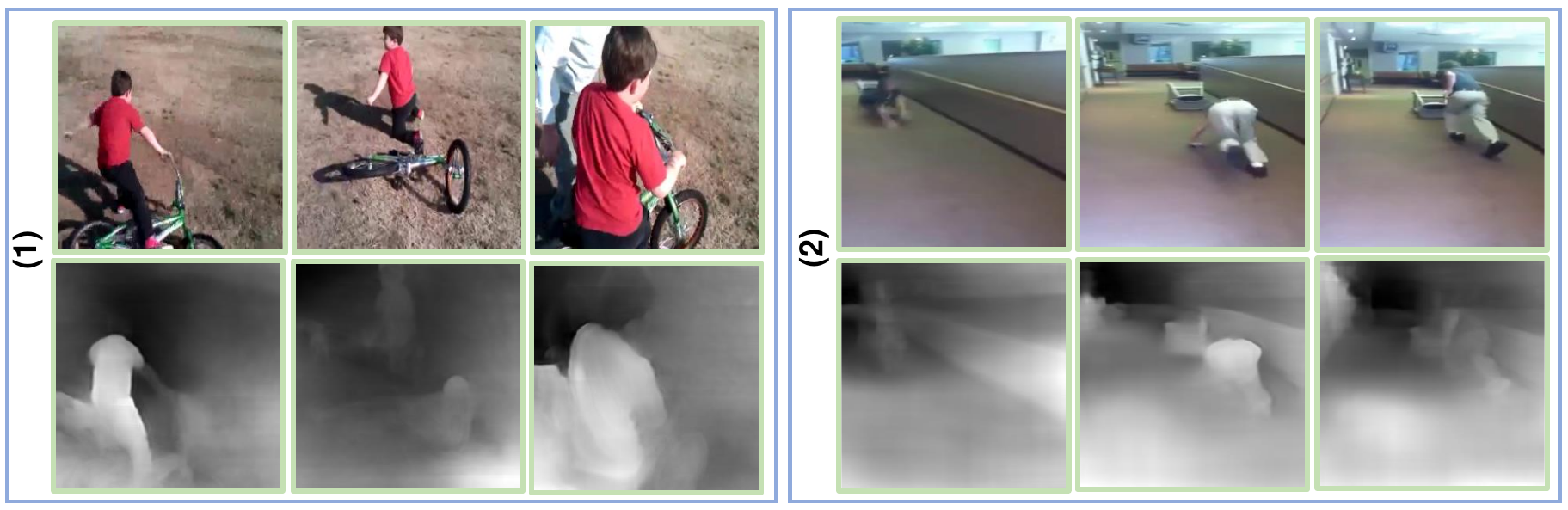}
        \caption{Two Correct Classification Cases: Here are two examples where the classification results of UMA-FD are correct. Each example contains three frames from the video. The first line is RGB data, and the second line is the corresponding Depth data.}
        \label{single_case_right}
\end{figure*}
\subsection{Ablation Study}
Next, we compare the individual contributions of different components of UMA-FD.
We use the backbone of X3D and gradually add the various module parts and loss functions of our proposed method based on the baseline method to conduct ablation experiments. We report the results in Table \ref{table_ablation}, which illustrates the contribution of the various building blocks and corresponding loss functions. For the convenience of description, we number each experiment: for example, the number of the baseline is V-01.

First, we add the modality head and modality loss to the baseline model, which is refered to as V-02.
The accuracy is increased from 73\% to 75.25\% compared to V-01, with an absolute value improvement of 2.25\%, and F1 score and AUC also improve significantly. The results show that the modality head and modality loss can indeed alleviate the difference between the features of different modalities, so that more information learned on RGB data can be applied to depth data, and this conclusion will be further confirmed by the qualitative results in next part.

Building upon the optimal model from V-02, we set a threshold to distinguish positive and negative sample pseudo-labels and incorporate a pseudo loss, resulting in the V-03 model. The model's accuracy has been further improved, with the optimal accuracy reaching 76.25\%, a 1\% increase compared to V-02. The F1 score has also significantly improved, and the AUC is slightly better than the previous version, demonstrating an overall noticeable improvement.
V-03 capitalizes on V-02 ability to classify depth data and employs partial pseudo-label information with higher confidence, enabling the model to learn more depth data information.

Next,  we verify the effect of the bridge feature loss, based on the optimal model of V-03. We add the IDM and bridge feature loss (V-04), leading to an accuracy gain from 76.25\% to 77.25\%.
In this version, F1 score drop significantly, and we believe this is due  to the large deviation between precision and recall caused by the suboptimal threshold of 0.5. Adjusting the threshold improves the outcome significantly:
the AUC is increased to 84.67\% compared to 83.05\% in V-03. Overall, the performance has been  improved, which shows that the IDM module and bridge feature loss indeed make the difference of representations of different modalities smaller.

Next, we add XBM\_triplet loss to V-04, and refer to it as V-05.
The XBM\_triplet loss also requires the labels of samples.
For the RGB sample data, the real label is used directly.
For the unlabeled depth data, the pseudo label that meets the threshold is used to calculate the triplet loss.
When looking at the optimal results of V-05 compared with V-04, the accuracy increase from 77.25\% to 77.75\%.
Comprehensive comparison of accuracy with F1 score and AUC, V-05 is also better than V-04, which 
shows that the XBM\_triplet loss is also effective for  cross-modal fall detection.

Finally, we verify  the effectiveness of the loss weight adaptation method. Based on the optimal model from V-05, we incorporate the loss weight adaptive network to automatically learn the weight of each loss. This model, numbered V-06, is our proposed method, UMA-FD. The accuracy has increased from 77.75\% to 78.5\%, representing an absolute value increase of 0.75\%. The F1 score and AUC are the best among all experimental results, showcasing a clear improvement. This demonstrates that the loss weight adaptive network can indeed find the optimal model more rapidly than manual parameter adjustment.

The above ablation experiments provide a comprehensive verification of each individual module. Finally, by employing cross-modal unsupervised adaptive learning, the classification accuracy of the unlabeled depth data is increased from 73\%  (baseline) to 78.5\%, with an impressive gain of 5.5\%.

\subsection{Qualitative Results}
We present the t-SNE visualization of the RGB data and depth data feature spaces $M(\cdot)$ generated by the baseline method, the method incorporating modality loss, and our proposed method UMA-FD in Fig. \ref{tSNE_analysis}. It is evident that our proposed method does mitigate the differences between the source and target modalities to some extent.

In the baseline method, we directly use RGB model to predict depth data, and the resulting feature distributions differ significantly.
When we add modality loss to eliminate these differences, the situation becomes much better.
In our final proposed method UMA-FD, the feature distributions are essentially mixed together, which is the desired outcome.
Therefore, we are able to more effectively utilize the information learned from the RGB data when applied to the depth data, thereby improving the classification accuracy of depth data. However, the feature space distributions of the two modalities' data still differ significantly, and even with our proposed method UMA-FD, these differences are not completely eliminated. Identifying ways to further reduce these differences is an important direction for future work in the proposed task.

Our next area of investigation is model interpretability.
In Fig. \ref{model_interpretability}, we showcase backbone GradCAM~\cite{selvaraju2017grad} results of the best baseline and UMA-FD model, which is generated using the  Python package~\cite{jacobgilpytorchcam}.
Each line represents  an example from the test set, and the four illustrations are GradCAM-based visualization of results of our baseline for RGB data, baseline for depth data, UMA-FD for RGB data, and UMA-FD for depth data (from left to right respectively).

For the two examples, the classification results of the baseline are incorrect, while the classification results of UMA-FD are accurate. For RGB data, both the baseline and UMA-FD can capture key areas in the video image. However, for depth data, the baseline fails to capture the key area, while UMA-FD successfully captures it.
In UMA-FD, the modality loss and the bridge feature loss help mitigate the differences in features between the modalities, while the pseudo-label loss and the XBM\_triplet loss allow the model to obtain more information from depth data. These factors enable our method to better capture key areas in depth data. The GradCAM results show that, compared to the baseline, our proposed method UMA-FD more effectively transfers the RGB information to the depth data, validating the effectiveness of cross-modal unsupervised learning and explaining why UMA-FD achieves higher accuracy.

Finally, we analyze the classification results of UMA-FD for multiple examples.
In Fig. \ref{single_case_error}, we provide two fall samples where the classification outcome is  wrong, while Fig. \ref{single_case_right} showcases two correctly assigned examples. 
Three frames of RGB images in a video sample and their corresponding depth frames are extracted. For the first case of incorrect prediction, the falling action is located in the corner of the image with a small magnitude, and it occupies only a few frames. For the second case of incorrect prediction, the person falling is too far from the camera, and their presence in the image takes up very few pixels.
For the two correct  cases, the distance and magnitude of the falling action are relatively normal. The above analysis demonstrates that our model still has limitations in recognizing falling actions with short durations and small magnitudes. Additionally, detecting falls in cases where the person is far from the camera remains challenging, as the model struggles to extract sufficient cues from these examples.
Addressing these two limitations will be valuable research directions in our future work.

\section{Conclusion and Future Work}

In this paper, we for the first time addressed  unsupervised modality adaptation (UMA) for fall detection, extending the concept of unsupervised domain adaptation (UDA) to accommodate specific application requirements. We generate a dual-modality fall detection dataset comprising RGB and depth data, containing 2979 samples and surpassing the size of most existing dual-modality sequence fall detection datasets. This dataset is created based on the public Kinetics-700 database and an off-the-shelf depth estimation algorithm. To enhance the classification accuracy of the unlabeled depth data, we apply a series of UDA methods to the UMA scenario and achieve scene adaptation.
In single-task multi-loss scenarios, manually adjusting the weight of each loss is time-consuming, labor-intensive, and unlikely to yield optimal results. To address this issue, we design a loss weight adaptive network that automatically learns the weight of each loss. By integrating these optimization methods, we significantly improve our model's performance, increasing the classification accuracy from 73\% (baseline) to 78.5\% on our generated kinetics-700 dataset, demonstrating the feasibility of cross-modal unsupervised adaptive learning.

For future work, we aim to further improve our model, as there is still a considerable gap between the current cross-modal accuracy and the supervised target method. This may involve developing new techniques to further mitigate feature differences between modalities or enhancing the model's ability to detect targets at long distances or with small magnitudes. Additionally, we plan to explore the application of UMA in other scenarios, investigating unsupervised modality adaptation across RGB, depth, and other data modalities, such as point cloud data, to tackle the problem of cross-modal
unsupervised adaptive learning that exists in reality.

\bibliography{uma_fall_detection}

\begin{IEEEbiography}[{\includegraphics[width=1in,height=1.25in,clip,keepaspectratio]{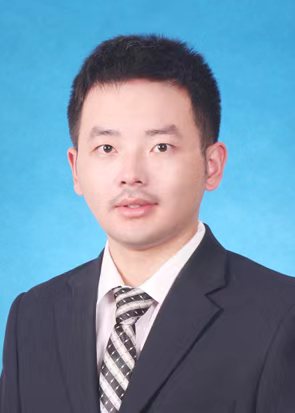}}]{Hejun Xiao} received the B.S. degree in computer science from Northeastern University, Shenyang, China, in 2017. And he received the M.S. degree in computer science and technology from Shanghai Jiao Tong University, Shanghai, China, in 2020. 
From 2020 to 2022, he worked as an algorithm engineer at Gaode, Alibaba, in China.
He is currently an Assistant Researcher in Xiong'an Institute of Innovation, Chinese Academy of Science, China. His research interests encompass 
computer vision, action recognition, fall detection and cognitive intelligence.
\end{IEEEbiography}

\begin{IEEEbiography}[{\includegraphics[width=1in,height=1.25in,clip,keepaspectratio]{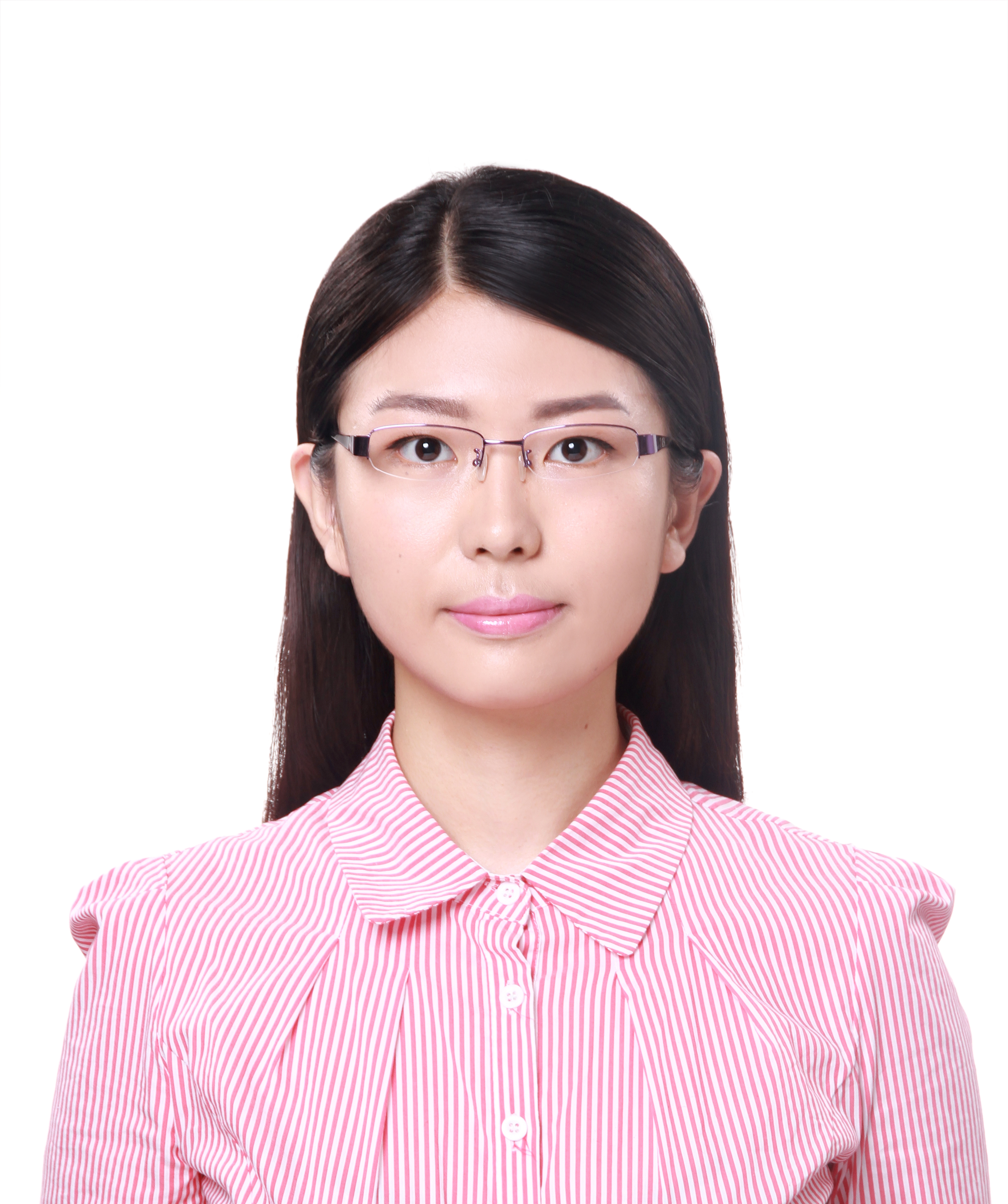}}]{Kunyu Peng} received her B.Sc. degree in Automation from Beijing Institute of Technology (BIT) in 2017. She received her M.Sc degree in Electrical Engineering and Information Technology from Karlsruhe Institute of Technology (KIT) in 2021 and is currently a research assistant and a Ph.D. candidate at the Computer Vision for Human-Computer Interaction Lab at KIT. She completed three internships separately at Chinese Academy of Science (Beijing, China), ESME Sudria (Paris, France) and Bosch (Leonberg, Germany). Her research fields include human activity recognition, scene understanding and intelligent vehicles.
\end{IEEEbiography}

\begin{IEEEbiography}[{\includegraphics[width=1in,height=1.25in,clip,keepaspectratio]{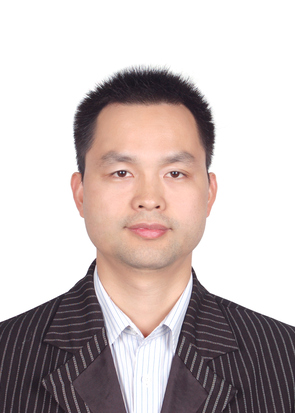}}]{Xiangsheng Huang} received the B.S. degree in
material science and the M.S. degree in computer science from Chongqing University, Chongqing, China, in 1998 and 2002, respectively, and the Ph.D. degree from the Institute of Automation, Chinese Academy of Sciences, Beijing, China, in 2005.
From 2005 to 2010, he was with the Samsung Advanced Institute of Technology. From 2010 to 2021, he was an Associate Professor with the Institute of Automation, Chinese Academy of Sciences. And He is currently a Professor with Xiong'an Institute of Innovation, Chinese Academy of Science.
His current research interests include cognitive intelligence, machine learning, cross-modal learning, 3-D imaging and registration, face technology and wavelet and filter banks.
\end{IEEEbiography}

\begin{IEEEbiography}[{\includegraphics[width=1in,height=1.25in]{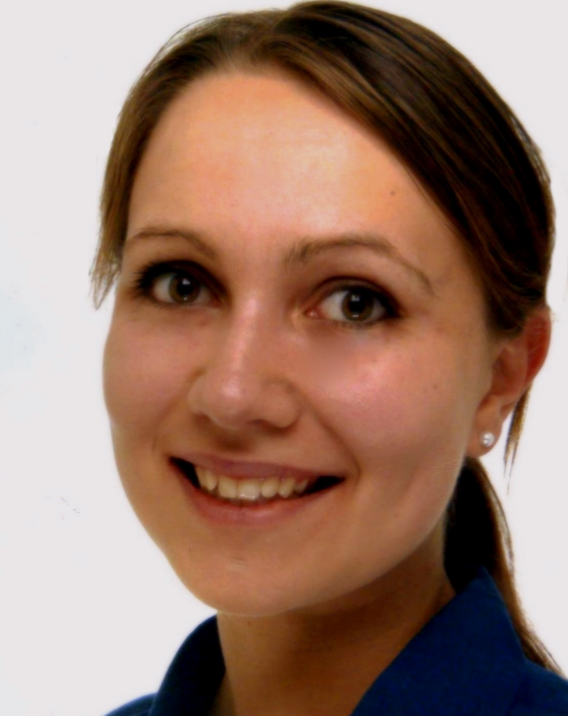}}]{Alina Roitberg}
received her B.Sc. and M.Sc. degrees in Computer Science with distinction from Technical University of Munich (TUM) in 2015. After working as a data science consultant in the automotive sector (2016-2017), she joined Karlsruhe Institute of Technology (KIT) for doctoral studies, during which she received the best student paper runner-up award at IV 2020 and completed a Ph.D. internship at Facebook.
She received her Ph.D. in deep learning for driver observation in 2021 and is currently a postdoctoral researcher at the Computer Vision for Human-Computer Interaction Lab at KIT.
Her research interests include human activity recognition, uncertainty-aware deep learning, open-set, zero- and few-shot recognition and applications in intelligent vehicles. 
\end{IEEEbiography}

\begin{IEEEbiography}[{\includegraphics[width=1in,height=1.25in,clip,keepaspectratio]{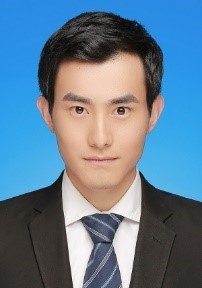}}]{Hao Li} received the M.S. degree in Geological engineering from China University of Geosciences (Beijing) ,China, in 2020. From 2018 to 2021, he worked as algorithm engineer in Beijing Yiyun Science and Technology Co., LTD., Geophysical Prospecting Institute of China General Administration of Metallurgical Geology. He is currently an assistant researcher at the Xiong'an Institute of Innovation, Chinese Academy of Sciences. His research interests include computer vision, millimeter wave radar imaging.
\end{IEEEbiography}

\begin{IEEEbiography}[{\includegraphics[width=1in,height=1.25in,clip,keepaspectratio]{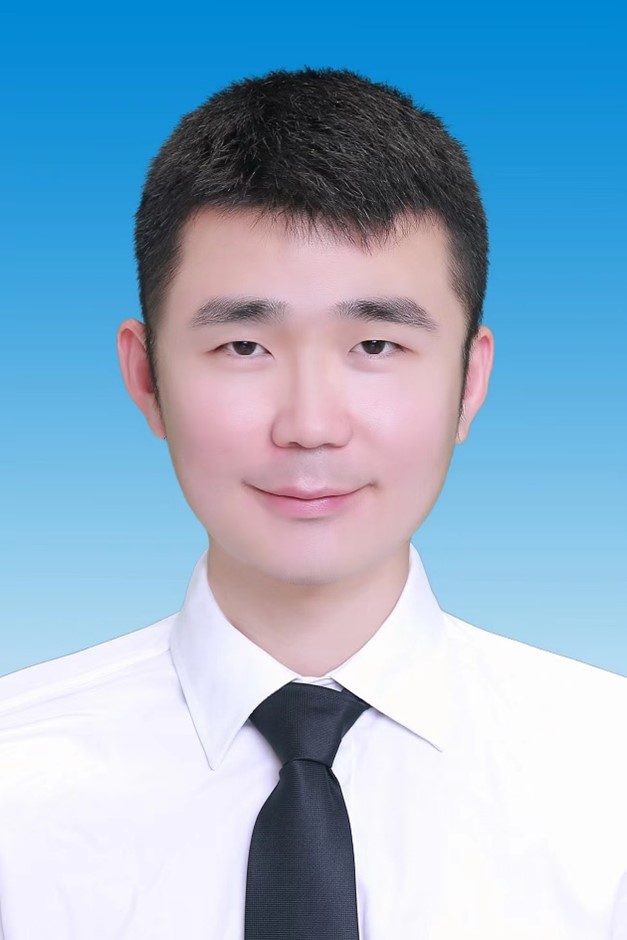}}]{Zhaohui Wang} received his B.S. degree from Xi’an University of Posts and Telecommunications. Since graduation, he has worked as an intern at Xiong'an Institute of Innovation, Chinese Academy of Science, China. He is now studying for a academic M.S. degree. His research interests include computer vision, object detection.
\end{IEEEbiography}

\begin{IEEEbiography}[{\includegraphics[width=1in,height=1.25in,clip,keepaspectratio]{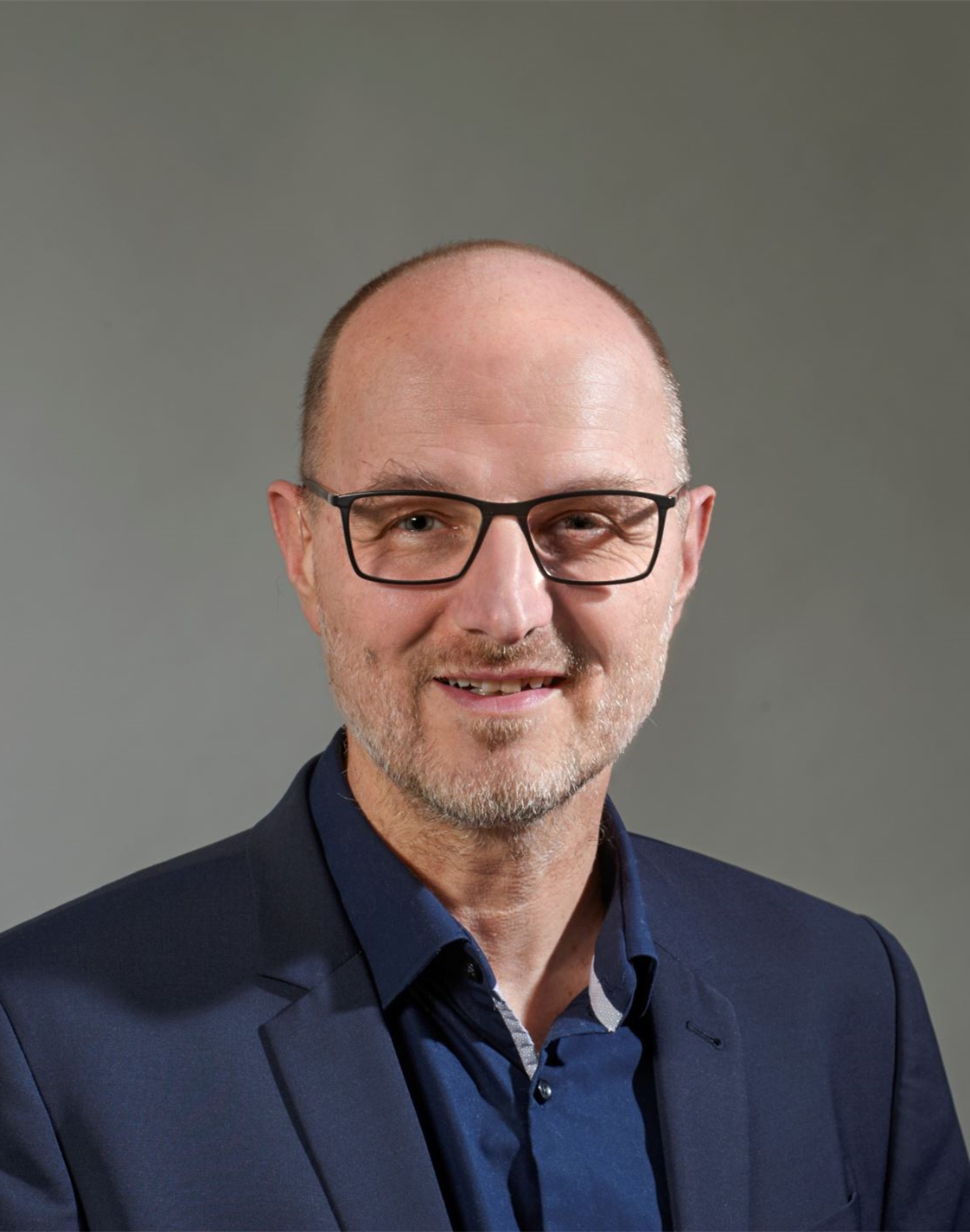}}]{Rainer Stiefelhagen} received his Diplom (Dipl.-Inform) and Doctoral degree (Dr.-Ing.) from the Universit\"at Karlsruhe (TH) in 1996 and 2002, respectively. He is currently a full professor for ``Information technology systems for visually impaired students'' at the Karlsruhe Institute of Technology (KIT), where he directs the Computer Vision for Human-Computer Interaction Lab at the Institute for Anthropomatics and Robotics as well as KIT’s Study Center for Visually Impaired Students. His research interests include computer vision methods for visual perception of humans and their activities, in order to facilitate perceptive multimodal interfaces, humanoid robots, smart environments, multimedia analysis and assistive technology for persons with visual impairments.
\end{IEEEbiography}
\end{document}